\documentclass[10pt,twocolumn,letterpaper]{article}

\usepackage{cvpr}
\usepackage{times}
\usepackage{epsfig}
\usepackage{graphicx}
\usepackage{amsmath}
\usepackage{amssymb}

\usepackage{xcolor}
\usepackage{booktabs}
\usepackage{multirow}
\usepackage{colortbl}
\usepackage{tabularx}

\definecolor{kuanfang}{RGB}{0, 0, 255}

\newcommand*{\affaddr}[1]{#1} 
\newcommand*{\affmark}[1][*]{\textsuperscript{#1}}

\usepackage[pagebackref=true,breaklinks=true,letterpaper=true,colorlinks,bookmarks=false]{hyperref}

\cvprfinalcopy 

\def\cvprPaperID{1218} 

\begin{document}

\title{Scene Memory Transformer for Embodied Agents in Long-Horizon Tasks}

\author{
Kuan Fang\affmark[1]
\and
Alexander Toshev\affmark[2]
\and
Li Fei-Fei\affmark[1]
\and
Silvio Savarese\affmark[1]
\and
\makebox[.5\linewidth]{
    \affaddr{\affmark[1]Stanford University}
    \qquad\qquad\qquad\qquad
    \affaddr{\affmark[2]Google Brain}
    }
}

\maketitle

\begin{abstract}
Many robotic applications require the agent to perform long-horizon tasks in partially observable environments. In such applications, decision making at any step can depend on observations received far in the past. Hence, being able to properly memorize and utilize the long-term history is crucial. In this work, we propose a novel memory-based policy, named Scene Memory Transformer (SMT). The proposed policy embeds and adds each observation to a memory and uses the attention mechanism to exploit spatio-temporal dependencies. This model is generic and can be efficiently trained with reinforcement learning over long episodes. On a range of visual navigation tasks, SMT demonstrates superior performance to existing reactive and memory-based policies by a margin. 

\end{abstract}

\section{Introduction}
\label{sec:introduction}


Autonomous agents, controlled by neural network policies and trained with reinforcement learning algorithms, have been used in a wide range of robot navigation applications~\cite{Anderson2018OnEO, mattersim2018cvpr, Arulkumaran2017ABS, kempka2016vizdoom, Mirowski2016LearningTN, Oh2016ControlOM, Xia2018GibsonER, Zhu2017VisualSP, Zhu2017TargetdrivenVN}. In many of these applications, the agent needs to perform tasks over long time horizons in unseen environments. Consider a robot patrolling or searching for an object in a large unexplored building. Typically, completing such tasks requires the robot to utilize the received observation at each step and to grow its knowledge of the environment, \eg building structures, object arrangements, explored area , \etc. Therefore, it is crucial for the agent to maintain a detailed memory of past observations and actions over the task execution.

\begin{figure}[!t]
    \centering
    \includegraphics[width=0.46\textwidth]{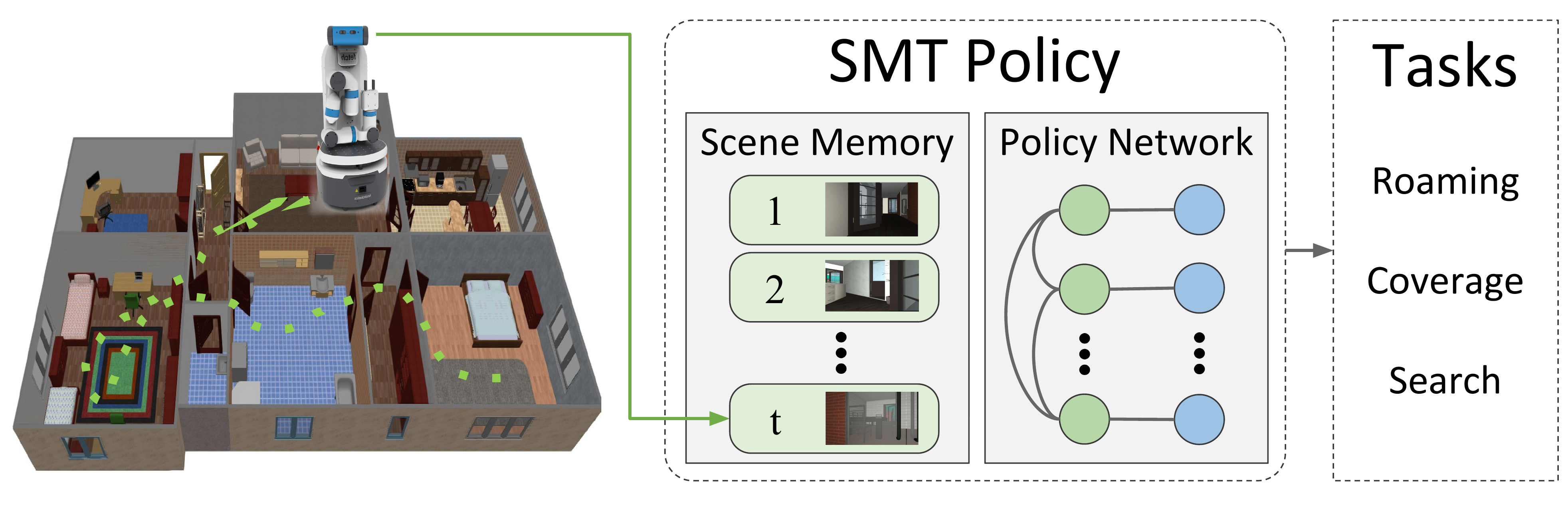}
    \caption{The Scene Memory Transformer (SMT) policy embeds and adds each observation to a memory. Given the current observation, the attention mechanism is applied over the memory to produce an action. SMT is demonstrated successfully in several visual navigation tasks, all of which has long time horizons.}
    \label{fig:intro}
    \vspace{-10pt}
\end{figure}


The most common way of endowing an agent's policy with memory is to use recurrent neural networks (RNNs), with LSTM~\cite{bakker2002reinforcement} as a popular choice. An RNN stores the information in a fixed-size state vector by combining the input observation with the state vector at each time step. The policy outputs actions for the agent to take given the updated state vector. Unfortunately, however, RNNs often fail to capture long-term dependencies \cite{Pascanu2013OnTD}. 

To enhance agent's ability to plan and reason, neural network policies with external memories have been proposed~\cite{Oh2016ControlOM,zhang2016learning}. Such memory-based policies have been primarily studied in the context of robot navigation  in partially observable environments, where the neural network learns to encode the received observations and write them into a map-like memory~\cite{Gupta2017CognitiveMA, Gupta2017UnifyingMA, Henriques2018MapNetA, Parisotto2017NeuralMS, Wierstra2010RecurrentPG}. Despite their superior performance compared to reactive and RNN policies, existing memory-based policies suffer from limited flexibility and scalability. Specifically, strong domain-specific inductive biases go into the design of such memories, \eg~2D layout of the environment, predefined size of this layout, geometry-based memory updates, \etc. Meanwhile, RNNs are usually critical components for these memory-based policies for exploiting spatio-temporal dependencies. Thus they still suffer from the drawbacks of RNN models.

In this work, we present \textit{Scene Memory Transformer (SMT)}, a memory-based policy using attention mechanisms, for understanding partially observable environments in long-horizon robot tasks. This policy is inspired by the Transformer model~\cite{Vaswani2017AttentionIA}, which has been successfully applied to multiple natural language processing problems recently. As shown in Fig.~\ref{fig:intro}, SMT consists of two modules: a scene memory which embeds and stores all encountered observations and a policy network which uses attention mechanism over the scene memory to produce an action. 

The proposed SMT policy is different from existing methods in terms of how to utilize observations received in the previous steps. Instead of combining past observations into a single state vector, as commonly done by RNN policies, SMT separately embeds the observations for each time step in the scene memory. In contrast to most existing memory models, the scene memory is simply a set of all embedded observations and any decisions of aggregating the stored information are deferred to a later point. We argue that this is a crucial property in long-horizon tasks where computation of action at a specific time step could depend on any provided information in the past, which might not be properly captured in a state vector or map-like memory. The policy network in SMT adopts attention mechanisms instead of RNNs to aggregate the visual and geometric information from the scene memory. This network efficiently learns to utilize the stored information and scales well with the time horizon. As a result, SMT effectively exploits long-term spatio-temporal dependencies without committing to a environment structure in the model design. 

Although the scene memory grows linearly with the length of the episode, it stores only an embedding vector at each steps. Therefore, we can easily store hundreds of observations without any burden in the device memory. This overhead is justified as it gives us higher performance compared to established policies with more compact memories.

Further, as the computational complexity of the original model grows quadratically with the size of the scene memory, we introduce a memory factorization procedure as part of SMT. This reduces the computational complexity to linear. The procedure is applied when the number of the stored observations is high. In this way, we can leverage a large memory capacity without the taxing computational overhead of the original model.

The advantages of the proposed SMT are empirically verified on three long-horizon visual navigation tasks: roaming, coverage and search.  We train the SMT policy using deep Q-learning~\cite{Mnih2015HumanlevelCT} and thus demonstrate for the first time how attention mechanisms introduced in \cite{Vaswani2017AttentionIA} can boost the task performance in a reinforcement learning setup. In these tasks, SMT considerably and consistently outperforms existing reactive and memory-based policies. Videos can found at \url{https://sites.google.com/view/scene-memory-transformer}
\section{Related Work}

\textbf{Memory-based policy using RNN.} Policies using RNNs have been extensively studied in reinforcement learning settings for robot navigation and other tasks. The most common architectural choice is LSTM~\cite{hochreiter1997long}. For example, Mirowski \textit{et al.}~\cite{Mirowski2016LearningTN} train an A3C~\cite{Mnih2016AsynchronousMF} agent to navigate in synthesized mazes with an LSTM policy. Wu \textit{et al.}~\cite{Wu2018BuildingGA} use a gated-LSTM policy with multi-modal inputs trained for room navigation. Moursavian \textit{et al.}~\cite{mousavian2018visual} use an LSTM policy for target driven navigation. The drawbacks of RNNs are mainly two-fold. First, merging all past observations into a single state vector of fixed size can easily lose useful information. Second, RNNs have optimization difficulties over long sequences~\cite{Pascanu2013OnTD, Trinh2018LearningLD} in backpropagation through time (BPTT). In contrast, our model stores each observations separately in the memory and only aggregate the information when computing an action. And it extracts spatio-temporal dependencies using attention mechanisms, thereby it is not handicapped by the challenges of BPTT. 


\textbf{External memory.} Memory models have been extensively studied in natural language processing for variety of tasks such as translation~\cite{Vaswani2017AttentionIA}, question answering~\cite{sukhbaatar2015end}, summarization~\cite{liu2018generating}. Such models are fairly generic, mostly based on attention functions and designed to deal with input data in the format of long sequences or large sets . 


When it comes to autonomous agents, most of the approaches structure the memory as a 2D grid. They are applied to visual navigation~\cite{Gupta2017CognitiveMA,Gupta2017UnifyingMA,Parisotto2017NeuralMS,zhang2017neural}, interactive question answering~\cite{gordon2017iqa}, and localization~\cite{chaplot2018active,Henriques2018MapNetA, jonschkowski2018differentiable}. These methods exhibit certain rigidity. For instance, the 2D layout is of fixed size and same amount of memory capacity is allocated to each part of the environment. Henriques \textit{et al.}~\cite{Henriques2018MapNetA} designs a differentiable mapping module with 2.5D representation of the spatial structure. Such a structured memory necessiates write operations, which compress all observations as the agent executes a task and potentially can lose information which could be useful later in the task execution. On the contrary, our SMT keeps all embedded observations and allows for the policy to attend to them as needed at any step. Further, the memory operations in the above papers are based on current estimate of robot localization, where the memory is being modified and how it is accessed. In contrast, we keep all pose information in its original form, thus allow for potentially more flexible use.

A more generic view on memory for autonomous agents has been less popular. Savinov \textit{et al.}~\cite{savinov2018semi} construct a topolgical map of the environment, and uses it for planning. Oh at al.~\cite{Oh2016ControlOM} use a single-layer attention decoder for control problems. However, the method relies on an LSTM as a memory controller, which comes with the challenges of backpropgation through time. Khan \textit{et al.}~\cite{khan2017memory} apply the very general Differentiable Neural Computer~\cite{graves2016hybrid} to control problems. While this approach is hard to optimize and is applied on very simple navigation tasks.

\begin{figure*}[!t]
    \centering
    \includegraphics[width=0.98\textwidth]{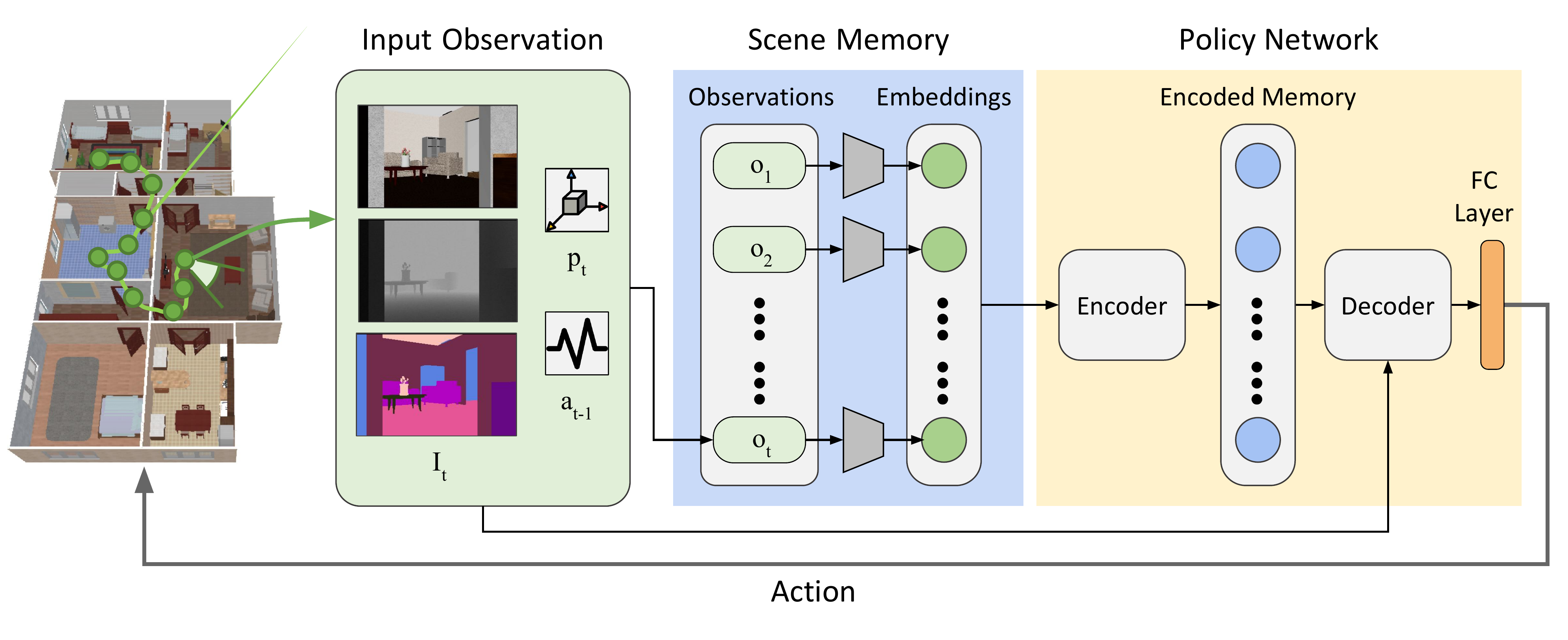}
    \caption{The \textbf{Scene Memory Transformer (SMT)} policy. At each time step $t$, the observation $\mathbf{o}_t$ is embedded and added to the scene memory. SMT has access to the full memory and produces an action according to the current observation. }
    \label{fig:model}
    \vspace{-5pt}
\end{figure*}

\textbf{Visual Navigation.} We apply SMT on a set of visual navigation tasks, which have a long history in computer vision and robotics~\cite{BoninFont2008VisualNF, desouza2002vision, thrun2007simultaneous}. Our approach falls into visual navigation, where the agent does not have any scene-specific information about the environment~\cite{Davison2003RealTimeSL, dayoub2013vision, Sim2006AutonomousVE, Tomono20063DOM, Wooden2006AGT}. As in recent works on end-to-end training policies for navigation tasks \cite{mattersim2018cvpr, kempka2016vizdoom, Mirowski2016LearningTN, Xia2018GibsonER, Zhu2017VisualSP, Zhu2017TargetdrivenVN}, our model does not need a map of the environment provided beforehand. While \cite{kempka2016vizdoom, Mirowski2016LearningTN} evaluates their models in 3D mazes, our model can handle more structured environments as realistic cluttered indoor scenes composed of multiple rooms. In contrast to \cite{Mirowski2016LearningTN, Zhu2017VisualSP, Zhu2017TargetdrivenVN} which train the policy for one or several known scenes, our trained model can generalize to unseen houses.

\section{Method}
\label{sec:method}

In this section, we first describe the problem setup. Then we introduce the Scene Memory Transformer (SMT) and its variations as shown in Fig.~\ref{fig:model}.

\subsection{Problem Setup}
\label{sec:problem_setup}

We are interested in a variety of tasks which require an embodied agent to navigate in unseen environments to achieve the task goal. These tasks can be formulated as the Partially Observable Markov Decision Process (POMDP)~\cite{Kaelbling1998PlanningAA} $(\mathcal{S},\mathcal{A}, \mathcal{O}, R(s, a), T(s'|s, a), P(o | s))$ where $\mathcal{S}$, $\mathcal{A}$, $\mathcal{O}$ are state, action and observation spaces, $R(s, a)$ is the reward function, $T(s'|s, a)$ and $P(o | s)$ are transition and observation probabilities.

The observation is a tuple $o = (I, p, a_\textrm{prev}) \in \mathcal{O}$ composed of multiple modalities. $I$ represents the visual data consisting of an RGB image, a depth image and a semantic segmentation mask obtained from a camera sensor mounted on the robot. $p$ is the agent pose \wrt. the starting pose of the episode, estimated or given by the environment. $a_\textrm{prev}$ is the action taken at the previous time step. 

In our setup, we adopt a discrete action space defined as $\mathcal{A}=\{\mathtt{go\_forward}, \mathtt{turn\_left}, \mathtt{turn\_right}\}$, a common choice for navigation problems operating on a flat surface. Note that these actions are executed under noisy dynamics modeled by $P(s'|s, a)$, so the state space is continuous. 

While we share the same $\mathcal{O}$ and $\mathcal{A}$ across tasks and environments, each task is defined by a different reward function $R(s, a)$ as described in Sec.~\ref{sec:task_design}. The policy for each task is trained to maximize the expected return, defined as the cumulative reward $\mathbb{E}_\tau[\sum_t R(s_t, a_t)]$ over trajectories $\tau={(s_t, a_t)_{t=1}^{H}}$ of time horizon $H$ unrolled by the policy.

\subsection{Scene Memory Transformer}

The SMT policy, as outlined in Fig.~\ref{fig:model}, consists of two modules. The first module is the \textit{scene memory} $M$ which stores all past observations in an embedded form. This memory is updated at each time step. The second module, denoted by $\pi(a | o, M)$, is an attention-based \textit{policy network} that uses the updated scene memory to compute an distribution over actions.

In a nutshell, the model and its interaction with the environment at time $t$ can be summarized as:
\begin{eqnarray*}
o_{t} & \sim & P(o_{t} | s_t) \\
M_{t} & =& \textrm{Update}(M_{t-1}, o_t) \\
a_t & \sim & \pi(a_t | o_t, M_t) \\
s_{t+1} & \sim & T(s_{t+1} | s_t, a_t) 
\end{eqnarray*}
In the following we define the above modules.

\subsubsection{Scene Memory}
The scene memory $M$ is intended to store all past observations in an embedded form. It is our intent not to endow it with any geometric structure, but to keep it as generic as possible. Moreover, we would like to avoid any loss of information when writing to $M$ and provide the policy with all available information from the history. So we separately keep observations of each step in the memory instead of merging them into a single state vector as in an RNN.

The scene memory can be defined recursively as follows. Initially it is set to the empty set $\varnothing$. At the current step, given an observation $o = (I, p, a_\textrm{prev})$, as defined in Sec.~\ref{sec:problem_setup}, we first embed all observation modalities, concatenate them, and apply a fully-connected layer $\textrm{FC}$:
\begin{equation}
\psi(o) = \textrm{FC}(\{\phi_I(I), \phi_p(p), \phi_a(a_\textrm{prev})\})
\end{equation}
where $\phi_I$, $\phi_p$, $\phi_a$ are embedding networks for each modality as defined in Sec.~\ref{sec:implementation}. To obtain the memory for the next step, we update it by adding $\psi(o)$ to the set:
\begin{equation}
\textrm{Update}(M, o) = M \cup \{ \psi(o) \}
\end{equation}

The above memory grows linearly with the episode length. As each received observation is embedded into low-dimensional vectors in our design, one can easily store hundreds of time steps on the hardware devices. While RNNs are restricted to a fixed-size state vector, which usually can only capture short-term dependencies. 

\subsubsection{Attention-based Policy Network}\label{sec:smt}
\begin{figure}[!t]
    \centering
    \includegraphics[width=0.45\textwidth]{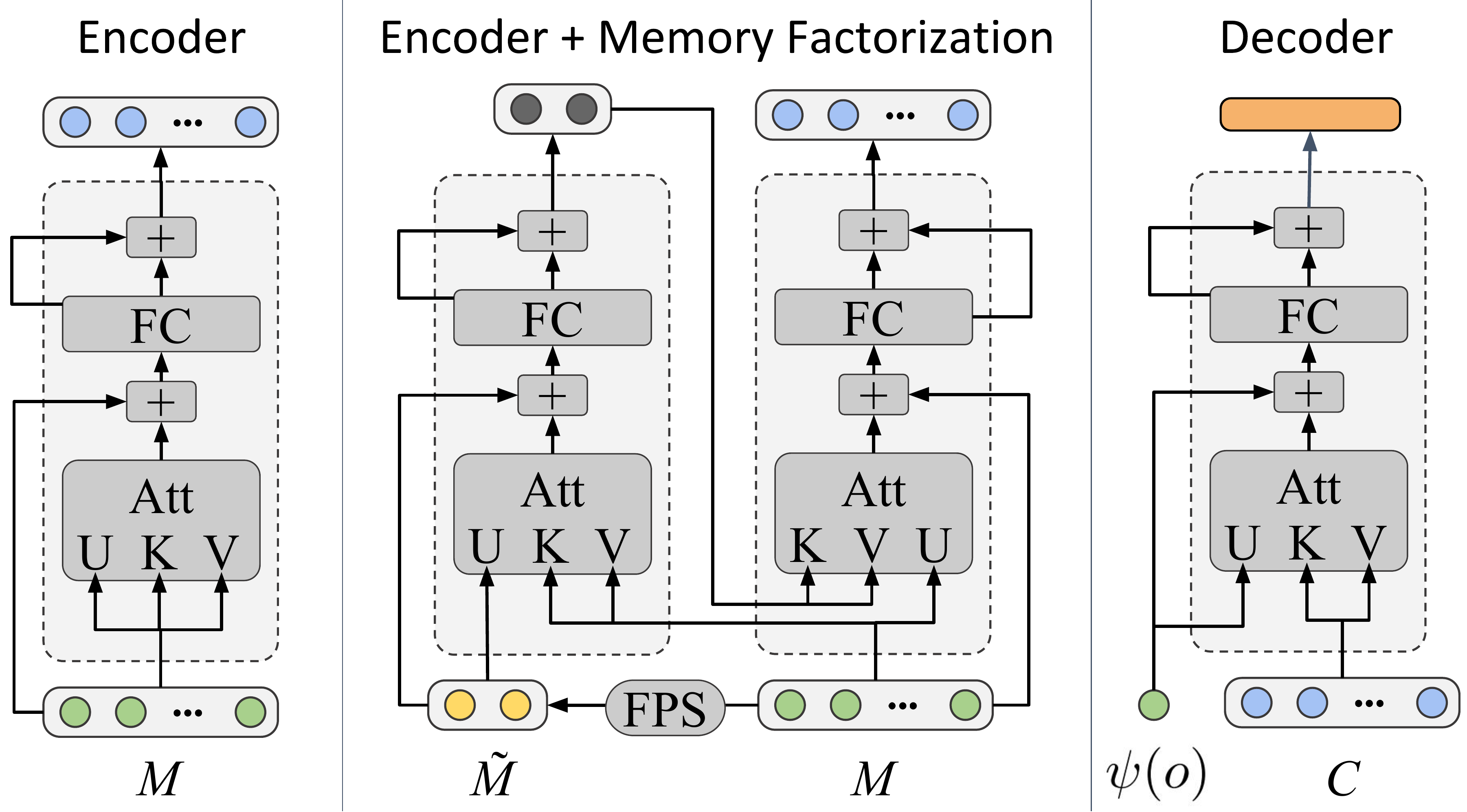}
    \caption{Encoder without memory factorization, encoder with memory factorization, and decoder as in Sec.~\ref{sec:smt}.}
    \label{fig:architectures}
    \vspace{-10pt}
\end{figure}

The policy network $\pi(a | o, M)$ uses the current observation and the scene memory to compute a distribution over the action space. As shown in Fig.~\ref{fig:model}, we first \textit{encode} the memory by transforming each memory element in the context of all other elements. This step has the potential to capture the spatio-temporal dependencies in the environment. Then, we \textit{decode} an action according to the current observation, using the encoded memory as the context.

\textbf{Attention Mechanism.} Both encoding and decoding are defined using attention mechanisms, as detailed by \cite{Vaswani2017AttentionIA}. In its general form, the attention function $\textrm{Att}$ applies $n_1$ attention queries $U\in\mathbb{R}^{n_1 \times d_k}$ over $n_2$ values $V\in\mathbb{R}^{n_2 \times d_v}$ with associated keys $K\in\mathbb{R}^{n_2 \times d_k}$, where $d_k$ and $d_v$ are dimensions of keys and values. The output of $\textrm{Att}$ has $n_1$ elements of dimension $d_v$, defined as a weighted sum of the values, where the weights are based on dot-product similarity between the queries and the keys:
\begin{equation}\label{eq:attention}
    \textrm{Att}(U, K, V) = \textrm{softmax}(UK^T) V
\end{equation}

An attention block $\textrm{AttBlock}$ is built upon the above function and takes two inputs $X\in\mathbb{R}^{n_1 \times d_x}$ and $Y\in\mathbb{R}^{n_2 \times d_y}$ of dimension $d_x$ and $d_y$ respectively. It projects $X$ to the queries and $Y$ to the key-value pairs. It consists of two residual layers. The first is applied to the above $\textrm{Att}$ and the second is applied to a fully-connected layer:
\begin{align}\label{eq:attention_block} 
\textrm{AttBlock}(X, Y) &= \textrm{LN}(\textrm{FC}(H) + H) \\
\textrm{where }H &= \textrm{LN}(\textrm{Att}(X W^U, Y W^K, Y W^V) + X) \nonumber
\end{align}
where $W^U\in\mathbb{R}^{d_x \times d_k}$, $W^K\in\mathbb{R}^{d_y \times d_k}$ and $W^V\in\mathbb{R}^{d_y \times d_v}$ are projection matrices and $\textrm{LN}$ stands for layer normalization~\cite{ba2016layer}. We choose $d_v = d_x$ for the residual layer.

\textbf{Encoder.} As in \cite{Vaswani2017AttentionIA}, our SMT model uses self-attention to encode the memory $M$. More specifically, we use $M$ as both inputs of the attention block. As shown in Fig.~\ref{fig:architectures}, this transforms each embedded observation by using its relations to other past observations:
\begin{equation}\label{eq:encoder}
    \textrm{Encoder}(M) = \textrm{AttBlock}(M, M)
\end{equation}
In this way, the model extracts the spatio-temporal dependencies in the memory.

\textbf{Decoder.} The decoder is supposed to produce actions based on the current observation given the context $C$, which in our model is the encoded memory. As shown in Fig.~\ref{fig:architectures}, it applies similar machinery as the encoder, with the notable difference that the query in the attention layer is the embedding of the current observation $\psi(o)$:
\begin{equation}\label{eq:decoder}
    \textrm{Decoder}(o, C) = \textrm{AttBlock}(\psi(o), C)
\end{equation}

The final SMT output is a probability distribution over the action space $\mathcal{A}$: 
\begin{eqnarray}
\pi(a|o, M) & = & \textrm{Cat}(\textrm{softmax}(Q)) \\
    \textrm{where }Q &=& \textrm{FC}(\textrm{FC}(\textrm{Decoder}(o,\textrm{Encoder}(M)))) \label{eq:qdef} \nonumber
\end{eqnarray}
where $\textrm{Cat}$ denotes categorical distribution.

This gives us a stochastic policy from which we can sample actions. Empirically, this leads to more stable behaviors, which avoids getting stuck in suboptimal states.

\textbf{Discussion.} The above SMT model is based on the encoder-decoder structure introduced in the Transformer model, which has seen successes on natural language processing (NLP) problems such as machine translation, text generation and summarization. The design principles of the model, supported by strong empirical results, transfer well from the NLP domain to the robot navigation setup, which is the primary motivation for adopting it.

First, an agent moving in a large environment has to work with dynamically growing number of past observations. The encoder-decoder structure has shown strong performance exactly in the regime of lengthy textual inputs. Second, contrary to common RNNs or other structured external memories, we do not impose a predefined order or structure on the memory. Instead, we encode temporal and spatial information as part of the observation and let the policy learn to interpret the task-relevant information through the attention mechanism of the encoder-decoder structure. 

\subsubsection{Memory Factorization} 
The computational complexity of the SMT is dominated by the number of query-key pairs in the attention mechanisms. Specifically, the time complexity is $O(|M|^2)$ for the encoder due to the self-attention, and $O(|M|)$ for the decoder. In long-horizon tasks, where the memory grows considerably, quadratic complexity can be prohibitive. Inspired by \cite{Lee2018SetT}, we replace the self-attention block from Eq.~(\ref{eq:attention_block}) with a composition of two blocks of similar design but more tractable computation:
\begin{equation}
    \textrm{AttFact}(M, \tilde M) = \textrm{AttBlock}(M, \textrm{AttBlock}(\tilde M, M))
\end{equation}
where we use a ``compressed'' memory $\tilde M$ obtained via finding representative centers from $M$. These centers need to be dynamically updated to maintain a good coverage of the stored observations. In practice, we can use any clustering algorithm. For the sake of efficiency, we apply iterative farthest point sampling (FPS)~\cite{qi2017pointnet++} to the embedded observations in $M$, in order to choose a subset of elements which are distant from each other in the feature space. The running time of FPS is in $O(|M||\tilde M|)$ and the final complexity of $\textrm{AttFact}$ is $O(|M||\tilde M|)$. With a fixed number of centers, the overall time complexity becomes linear. The diagram of the encoder with memory factorization is shown in Fig.~\ref{fig:architectures}.

\subsection{Training}

We train all model variants and baselines using the standard deep Q-learning algorithm \cite{Mnih2015HumanlevelCT}. We follow \cite{Mnih2015HumanlevelCT} in the use of an experience replay buffer, which has a capacity of 1000 episodes. The replay buffer is initially filled with episodes collected by a random policy and is updated every 500 training iterations. The update replaces the oldest episode in the buffer with a new episode collected by the updated policy. At every training iteration, we construct a batch of 64 episodes randomly sampled from the replay buffer. The model is trained with Adam Optimizer~\cite{kingma2014adam} with a learning rate of $5 \times 10^{-4}$. All model parameters except for the embedding networks are trained end-to-end. During training, we continuously evaluate the updated policy on the validation set (as in Sec.~\ref{sec:task_design}). We keep training each model until we observe no improvement on the validation set. 

The embedding networks are pre-trained using the SMT policy with the same training setup, with the only difference that the memory size is set to be 1. This leads to a SMT with no attention layers, as attention of size 1 is an identity mapping. In this way, the optimization is made easier so that the embedding networks can be trained end-to-end. After being trained to convergence, the parameters of the embedding networks are frozen for other models.

A major difference to RNN policies or other memory-based policies is that SMT does not need backpropagation through time (BPTT). As a result, the optimization is more stable and less computationally heavy. This enables training the model to exploit longer temporal dependencies.

\subsection{Implementation Details}\label{sec:implementation}

Image modalities are rendered as $640 \times 480$ and subsampled by 10. Each image modality is embedded into 64-dimensional vectors using a modified ResNet-18~\cite{he2016deep}. We reduce the numbers of filters of all convolutional layers by a factor of $4$ and use stride of 1 for the first two convolutional layers. We remove the global pooling to better capture the spatial information and directly apply the fully-connected layer at the end. Both pose and action vectors are embedded using a single $16$-dimensional fully-connected layer. 

Attention blocks in SMT use multi-head attention mechanisms~\cite{Vaswani2017AttentionIA} with 8 heads. The keys and values are both $128$-dimensional. All the fully connected layers in the attention blocks are $128$-dimensional and use ReLU non-linearity.

A special caution is to be taken with the pose vector. First, at every time step all pose vectors in the memory are transformed to be in the coordinate system defined by the current agent pose. This is consistent with an ego-centric representation of the memory. Thus, the pose observations need to be re-embedded at every time step, while all the other observations are embedded once. Second, a pose vector $p=(x, y, \theta)$ at time $t$ is converted to a normalized version $p=(x/\lambda, y/\lambda, \cos\theta, \sin\theta, e^{-t})$, embedding in addition its temporal information $t$ in a soft way in its last dimension. This allows the model to differentiate between recent and old observation, assuming that former could be more important than latter. The scaling factor $\lambda=5$ is used to reduce the magnitude of the coordinates. 
\section{Experiments}
\label{sec:experiments}

We design our experiments to investigate the following topics: 1) How well does SMT perform on different long-horizon robot tasks 2) How important is its design properties compared to related methods? 3) Qualitatively, what agent behaviors does SMT learn?
\subsection{Task Setup}
\label{sec:task_design}

To answer these questions, we consider three visual navigation tasks: \textit{roaming}, \textit{coverage}, and \textit{search}. These tasks require the agent to summarize spatial and semantic information of the environment across long time horizons. All tasks share the same POMDP from Sec.~\ref{sec:problem_setup} except that the reward functions are defined differently in each task.

\textbf{Roaming:} The agent attempts to move forward as much as possible without colliding. In this basic navigation task, a memory should help the agent avoid cluttered areas and oscillating behaviors. The reward is defined as $R(s, a)=1$ iff $a=\mathtt{go\_forward}$ and no collision occurs.

\textbf{Coverage:} In many real-world application a robot needs to explore unknown environments and visit all areas of these environments. This task clearly requires a detailed memory as the robot is supposed to remember all places it has visited. To define the coverage task, we overlay a grid of cell size $0.5$ over the floorplan of each environment. We would like the agent to visit as many unoccupied cells as possible, expressed by reward $R(s, a)=5$ iff robot entered unvisited cell after executing the action.

\textbf{Search:} To evaluate whether the policy can learn beyond knowledge about the geometry of the environment, we define a semantic version of the coverage tasks. In particular, for six target object classes\footnote{We use television, refrigerator, bookshelf, table, sofa, and bed.}, we want the robot to search for as many as possible of them in the house. Each house contains 1 to 6 target object classes, 4.9 classes in average. Specifically, an object is marked as found if more than $4\%$ of pixels in an image has the object label (as in \cite{Wu2018BuildingGA}) and the corresponding depth values are less than $2$ meter. Thus, $R(s, a)=100$ iff after taking action $a$ we find one of the six object classes which hasn't been found yet.

We add a collision reward of $-1$ for each time the agent collides. An episode will be terminated if the agent runs into more than $50$ collisions. To encourage exploration, we add coverage reward to the search task with a weight of $0.2$.

The above tasks are listed in ascending order of complexity. The coverage and search tasks are studied in robotics, however, primarily in explored environments and are concerned about optimal path planning~\cite{galceran2013survey}.

\textbf{Environment.} We use SUNCG~\cite{song2016ssc}, a set of synthetic but visually realistic buildings. We use the same data split as chosen by \cite{Wu2018BuildingGA} and remove the houses with artifacts, which gives us $195$ training houses and $46$ testing houses. We hold out 20\% of the training houses as a validation set for ablation experiments. We run 10 episodes in each house with a fixed random seed during testing and validation. The agent moves by a constant step size of $0.25$ meters with $\mathtt{go\_forward}$. It turns by $45^\circ$ degree in place with $\mathtt{turn\_left}$ or $\mathtt{turn\_right}$. Gaussian noise is added to simulate randomness in real-world dynamics. 

\textbf{Model Variants.} To investigate the effect of different model aspects, we conduct experiments with three variants: \textbf{SMT}, \textbf{SMT + Factorization}, and \textbf{SM + Pooling}. The second model applies SMT with $\textrm{AttFact}$ instead of $\textrm{AttBlock}$. Inspired by \cite{Eslami2018NeuralSR}, the last model directly applies a max pooling over the elements in the scene memory instead of using the encoder-decoder structure of SMT.

\textbf{Baselines.} We use the following baselines for comparison. A \textbf{Random} policy uniformly samples one of the three actions. A \textbf{Reactive} policy is trained to directly compute Q values using a purely feedforward net. It is two fully-connected layers on top of the embedded observation at every step. A \textbf{LSTM} policy~\cite{Mirowski2016LearningTN} is the most common memory-based policy. A model with arguably larger capacity, called \textbf{FRMQN}~\cite{Oh2016ControlOM}, maintains embedded observations in a fixed-sized memory, similarly as SMT. Instead of using the encode-decoder structure to exploit the memory, it uses an LSTM, whose input is current observation and output is used to attend over the memory. 

For all methods, we use the same pretrained embedding networks and two fully-connected layers to compute Q values. We also use the same batch size of 64 during training. To train LSTM and FRMQN, we use truncated back propagation through time of 128 steps.

\subsection{Comparative Evaluation}\label{sec:eval}

\begin{table}[!t]
\small
\begin{tabular*}{0.475\textwidth}{l@{\extracolsep{\fill}}lll}
\toprule
 Method &  Reward &  Distance &  Collisions\\
 \midrule
 Random & 58.3 & 25.3 & 42.7 \\
 Reactive \cite{Mirowski2016LearningTN} & 308.9 & 84.6 & 29.3 \\
 LSTM \cite{Mirowski2016LearningTN} & 379.7 & 97.9 & \textbf{11.4} \\
 FRMQN \cite{Oh2016ControlOM} & 384.2 & 99.5 & 13.8 \\
 \hline
 SM + Pooling & 366.8 & 96.7 & 20.1 \\
 SMT + Factorization & 376.4 & 98.6 & 17.9 \\
 SMT & \textbf{394.7} & \textbf{102.1} & 13.6 \\
 \bottomrule
\end{tabular*}
\caption{\textbf{Performance on Roaming.} The average of cumulative reward, roaming distance and number of collisions are listed. }
\label{table:roaming}
\end{table}

\begin{table}[!t]
\small
\begin{tabular*}{0.475\textwidth}{l@{\extracolsep{\fill}}ll}
\toprule
 Method &  Reward & Covered Cells \\
 \midrule
 Random & 94.2 & 27.4 \\
 Reactive \cite{Mirowski2016LearningTN} & 416.2 & 86.9 \\
 LSTM \cite{Mirowski2016LearningTN} & 418.1 & 87.8 \\
 FRMQN \cite{Oh2016ControlOM} & 397.7 & 83.2 \\
 \hline
 SM + Pooling & 443.9 & 91.5 \\
 SMT + Factorization & 450.1 & 99.3 \\
 SMT  & \textbf{474.6} & \textbf{102.5} \\
 \bottomrule
\end{tabular*}
\caption{\textbf{Performance on Coverage.} The average of cumulative reward and number of covered cells are listed. }
\label{table:coverage}
\end{table}

\begin{table}[!t]
\small
\begin{tabular*}{0.475\textwidth}{l@{\extracolsep{\fill}}lll}
\toprule
 Method & Reward & Classes & Ratio\\
 \midrule
 Random & 140.5 & 1.79 & 36.3\% \\
 Reactive \cite{Mirowski2016LearningTN} & 358.2 & 3.14 & 61.9\% \\
 LSTM \cite{Mirowski2016LearningTN} & 339.4 & 3.07 & 62.6\% \\
 FRMQN \cite{Oh2016ControlOM} & 411.2 & 3.53 & 70.2\% \\
 \hline
 SM + Pooling & 332.5 & 2.98 & 60.6\% \\
 SMT + Factorization & \textbf{432.6} & \textbf{3.69} & \textbf{75.0\%} \\
 SMT & 428.4 & 3.65 & 74.2\% \\
 \bottomrule
\end{tabular*}
\caption{\textbf{Performance on Search.} The cumulative of total reward, number of found classes and ratio of found classes are listed.}
\label{table:search}
\end{table}


The methods are compared across the three different tasks: roaming in Table~\ref{table:roaming}, coverage in Table~\ref{table:coverage}, and search in Table~\ref{table:search}. For each task and method we show the attained reward and task specific metrics. 

\textbf{Effect of memory designs.} Across all tasks, SMT outperforms all other memory-based models. The relative performance gain compared to other approaches is most significant for coverage (14\% improvements) and considerable for search (5\% improvements). This is consistent with the notion that for coverage and search memorizing all past observations is more vital. On roaming, larger memory capacity (in SMT case) helps, however, all memory-based approaches perform in the same ballpark. This is reasonable in the sense that maintaining a straight collision free trajectory is a relatively short-sight task.

In addition to memory capacity, memory access via attention brings improvements. For all tasks SMT outperforms SM + Pooling. The gap is particularly large for object search (Table~\ref{table:search}), where the task has an additional semantic complexity of finding objects. Similarly, having multi-headed attention and residual layers brings an improvement over a basic attention, as employed by FRMQN, which is demonstrated on both coverage and search.

The proposed memory factorization brings computational benefits, at no or limited performance loss. Even if it causes drop sometimes, the reward is better than SM + Pooling and other baseline methods.  

\textbf{Implications of memory for navigation.} It is also important to understand how memory aids us at solving navigation tasks. For this purpose, in addition to reward, we report number of covered cells (Table~\ref{table:coverage}) and number of found objects (Table~\ref{table:search}). For both tasks, a reactive policy presents a strong baseline, which we suspect learns general exploration principles. Adding memory via SMT helps boost the coverage by $18\%$ over reactive, and $17\%$ over LSTM policies and $23\%$ over simpler memory mechanism (FRMQN). We also observe considerable boosts of number of found objects by $5\%$ in the search task.

\begin{figure}[!t]
    \centering
    \includegraphics[width=0.45\textwidth]{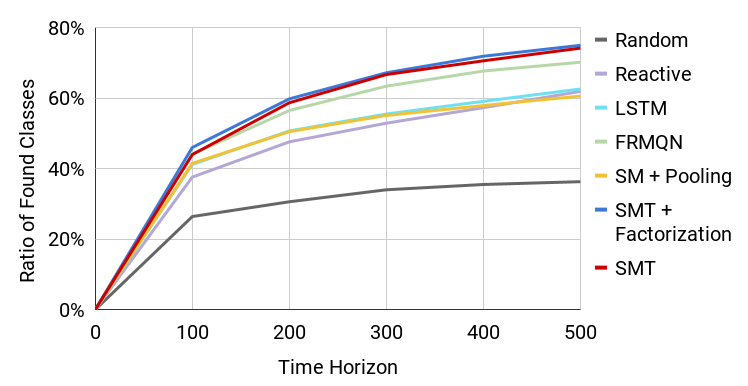}
    \caption{\textbf{Found classes by time steps.} For the search task, we show number of found target object classes across time steps. }
    \label{fig:search_breakdown}
    \vspace{-10pt}
\end{figure}

The reported metrics above are for a fixed time horizon of 500 steps. For varying time horizons, we show the performance on search in Fig.~\ref{fig:search_breakdown}. We see that memory-based policies with attention-based reads consistently find more object classes as they explore the environment, with SMT variants being the best. This is true across the full execution with performance gap increasing steadily up to 300 steps.

\begin{figure}[!t]
    \centering
    \includegraphics[width=0.47\textwidth]{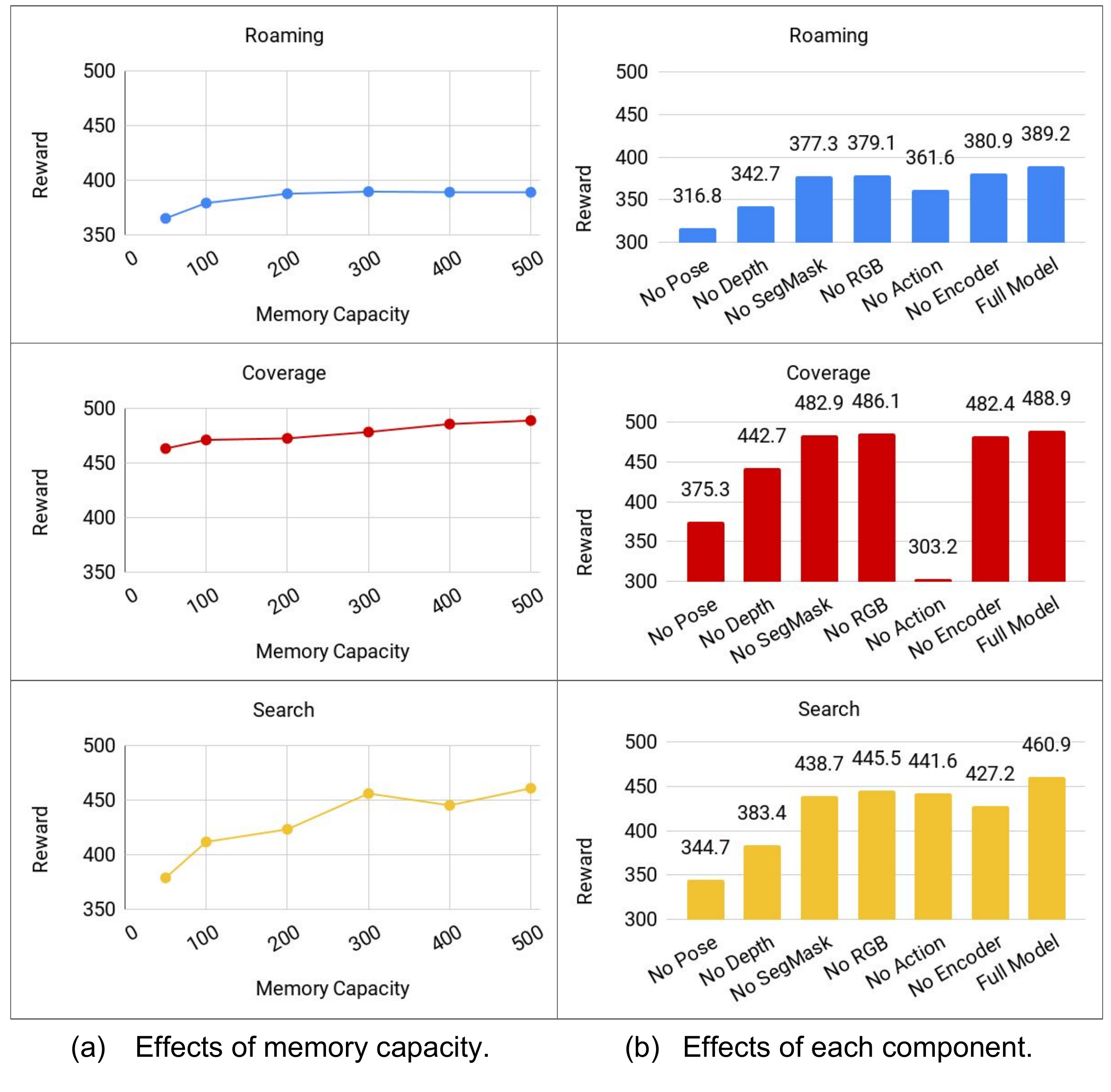}
    \caption{\textbf{Ablation Experiments.}  (a) We sweep the memory capacity from 50 steps to 500 steps and evaluate the reward of trajectories of 500 steps. (b) We leave out one component at a time in our full model and evaluate the averaged reward for each task.}
    \label{fig:ablation}
    \vspace{-10pt}
\end{figure}

\begin{figure*}[!t]
    \centering
    \includegraphics[width=\textwidth]{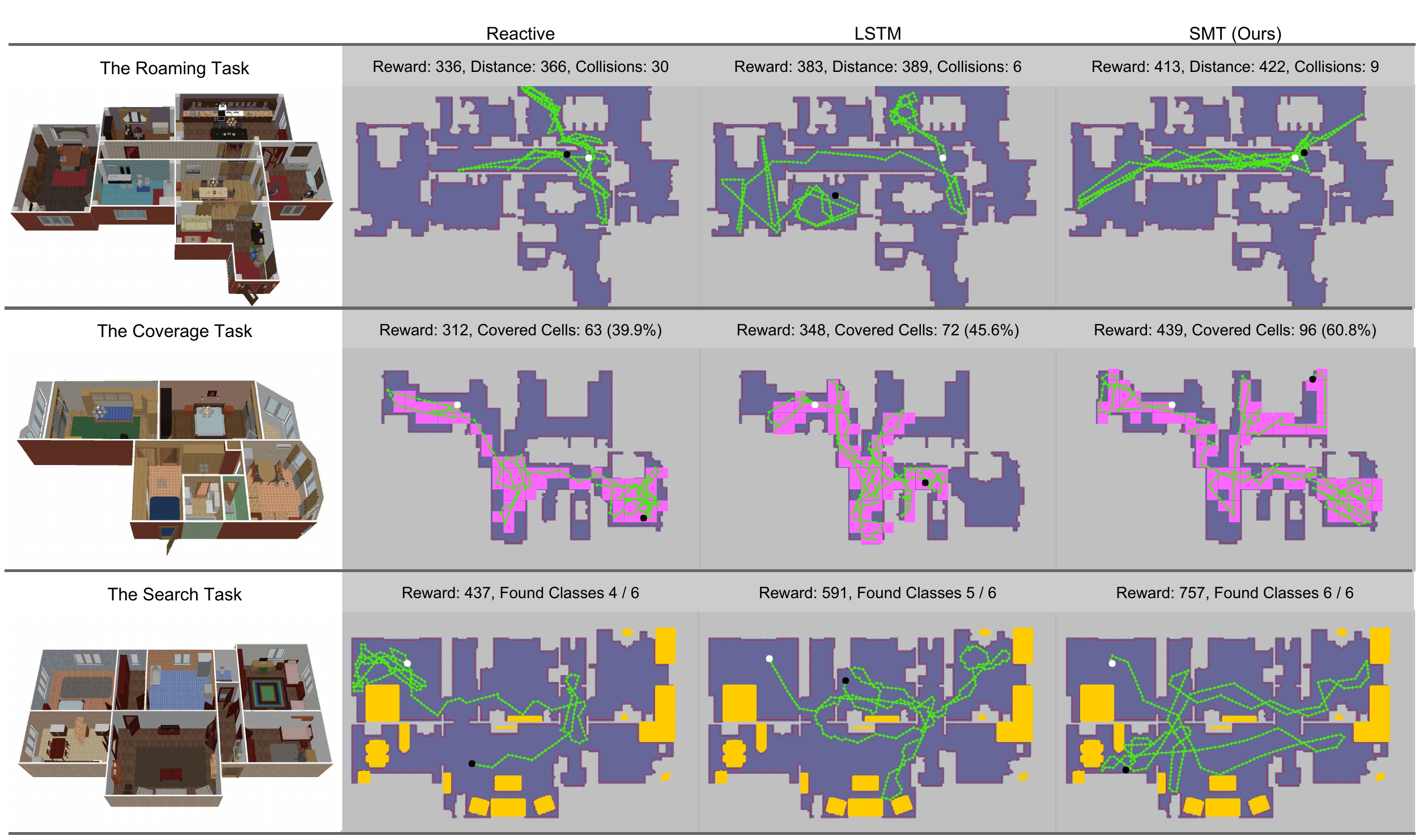}
    \caption{\textbf{Visualization of the agent behaviors.} We visualize the trajectories from the top-down view as green curves. Starting point and ending point of each trajectory are plot in white and black. Navigable area are masked in dark purple with red lines indicating the collision boundaries. For the coverage task, we mark the covered cells in pink. For the search task, we mark target objects with yellow masks.}
    \label{fig:qualitative}
    \vspace{-10pt}
\end{figure*}

\subsection{Ablation Analysis}
Here, we analyze two aspects of SMT: (i) size of the scene memory, and (ii) importance of the different observation modalities and componenets.

\textbf{Memory capacity.} While in the previous section we discussed memory capacity across models, here we look at the importance of memory size for SMT. Intuitively a memory-based policy is supposed to benefits more from larger memory over long time horizons. But in practice this depends on the task and the network capacity, as shown in Fig.~\ref{fig:ablation}~(a). All three tasks benefit from using larger scene memory. The performance of roaming grows for memory up to 300 elements. For coverage and search, the performance keeps improving constantly with larger memory capacities. This shows that SMT does leverage the provided memory. 

\textbf{Modalities and components.} For the presented tasks, we have image observations, pose, previous actions. To understand their importance, we re-train SMT by leaving out one modality at a time. We show the resulting reward in Fig.~\ref{fig:ablation}~(b). Among the observation modalities, last action, pose and the depth image play crucial roles across tasks. This is probably because SMT uses relative pose and last action to reason about spatial relationships. Further, depth image is the strongest clue related to collision avoidance, which is crucial for all tasks. Removing segmentation and RGB observations leads to little effect on coverage and drop of $10$ for roaming since these tasks are defined primarily by environment geometry. For search, however, where SMT needs to work with semantics, the drop is $15$ and $20$.

We also show that the encoder structure brings performance boost to the tasks. Especially in the search task, which is most challenging in terms of reasoning and planning, the encoder boosts the task reward by 23.7.


\subsection{Qualitative Results}
To better understand the learned behaviors of the agent, we visualize the navigation trajectories in Fig.~\ref{fig:qualitative}. We choose reactive and LSTM policies as representatives of memory-less and memory-based baselines to compare with SMT.

In the roaming task, our model demonstrates better strategies to keep moving and avoid collisions. In many of the cases, the agent first finds a long clear path in the house, which lets it go straight forward without frequently making turns. Then the agent navigates back and forth along the same route until the end of the episode. As a result, SMT usually leads to compact trajectories as shown in Fig.~\ref{fig:qualitative}, top row. In contrast, reactive policy and LSTM policy often wander around the scene with a less consistent pattern. 

In the coverage task, our model explores the unseen space more efficiently by memorizing regions that have been covered. As shown in Fig.~\ref{fig:qualitative}, middle row, after most of the cells inside a room being explored, the agent switches to the next unvisited room. Note that the cells are invisible to the agent, it needs to make this decision solely based on its memory and observation of the environment. It also remembers better which rooms have been visited so that it does not enter a room twice. 

In the search task, our model shows efficient exploration as well as effective strategies to find the target classes. The search task also requires the agent to explore rooms with the difference that the exploration is driven by target object classes. Therefore, after entering a new room, the agent quickly scans around the space instead of covering all the navigable regions. In Fig.~\ref{fig:qualitative}, if the agent finds the unseen target it goes straight towards it. Once it is done, it will leave the room directly. Comparing SMT with baselines, our trajectories are straight and direct between two targets, while baseline policies have more wandering patterns.
\vspace{-0.3cm}
\section{Conclusion}
This paper introduces Scene Memory Transformer, a memory-based policy to aggregate observation history in robotic tasks of long time horizons. We use attention mechanism to exploit spatio-temporal dependencies across past observations. The policy is trained on several visual navigation tasks using deep Q-learning. Evaluation shows that the resulting policy achieves higher performance to other established methods. 

\textbf{Acknowledgement:} We thank Anelia Angelova, Ashish Vaswani and Jakob Uszkoreit for constructive discussions. We thank Marek Fi\v{s}er for the software development of the simulation environment, Oscar Ramirez and Ayzaan Wahid for the support of the learning infrastructure.


\newpage
\appendix
\section{Environment Details}

In all experiments, we simulate a mobile base of the Fetch robot. The Fetch robot receives visual observations from a Primesense Carmine 1.09 short-range RGBD sensor mounted on its head. Accordingly, we render images of $640 \times 480$ resolution. To simulate the operation range of the depth sensor, we only render depth values for points that are within $5$ meters from the camera. We also provide a binary mask indicating which pixels have valid depth values and concatenate the mask with the depth image as its second channel. We also add zero-mean Gaussian noise of with a standard deviation of 0.05 meter to each pixel. The segmentation mask uses the class labels from NYU40~\cite{Eigen2015PredictingDS} with each pixel label encoded in the one-hot manner. We subsample the rendered images by a factor of $10$, providing us RGB images of $64 \times 48 \times 3$, depth images of $64 \times 48 \times 2$ and segmentation masks of $64 \times 48 \times 40$.

The environment dynamics is simulated for the Fetch robot operating on a planar surface. The robot moves forward and take turns by controlling the velocity of its two wheels, with a wheel radius of 0.065 meters and axis width of 0.375 meters. We add a zero-mean Gaussian with a standard deviation of 0.5 rad/s to both wheels to simulate the noisy dynamics. We check the collisions between the robot and the meshes of the environment. The robot will be reset to the previous pose when it collides by taking the action.
\section{Analysis of Memory Factorization}

In memory factorization, it is crucial to choose representative centers that have a good coverage of all past observations. Therefore, the centers should be distant from each other in the feature space. Since the memory keeps growing across time, the centers are supposed to be dynamically updated during the task execution instead of remaining as static vectors for all episodes.

In this section, we compare the \textbf{farthest point sampling (FPS)} used in SMT with two alternative types of representative centers. We refer to \textbf{Window} as the baseline which uses the last $|\tilde M|$ time steps in a fixed time window as representative centers. In this way, the centers are dynamically updated but only focus on the most recent history. We also implemented the static inducing points in \cite{Lee2018SetT}, which we refer to as \textbf{Static}. The  $|\tilde M|$ inducing points are trained as neural network weights and remain static during test time. We compare the performance of the three types of centers on the validation set by setting $|\tilde M|$ to be 100. As shown in Table.~\ref{table:centers}, FPS achieves comparable task performance with Static in the roaming task. And it outperforms the two baselines in coverage and search.

\begin{table}[!t]
\small
\begin{tabular*}{0.475\textwidth}{l@{\extracolsep{\fill}}lll}
\toprule
 Center Type & Roaming & Coverage & Search\\
 \midrule
 Window & 378.0 & 451.6 & 438.7 \\
 Static & \textbf{383.9} & 457.96 & 445.9 \\
 FPS & 383.3 & \textbf{481.2} & \textbf{462.7} \\
 \bottomrule
\end{tabular*}
\caption{Performance of using different types of representative center in memory factorization. Average rewards are listed. }
\label{table:centers}
\end{table}

\begin{figure}[!t]
    \centering
    \includegraphics[width=0.48\textwidth]{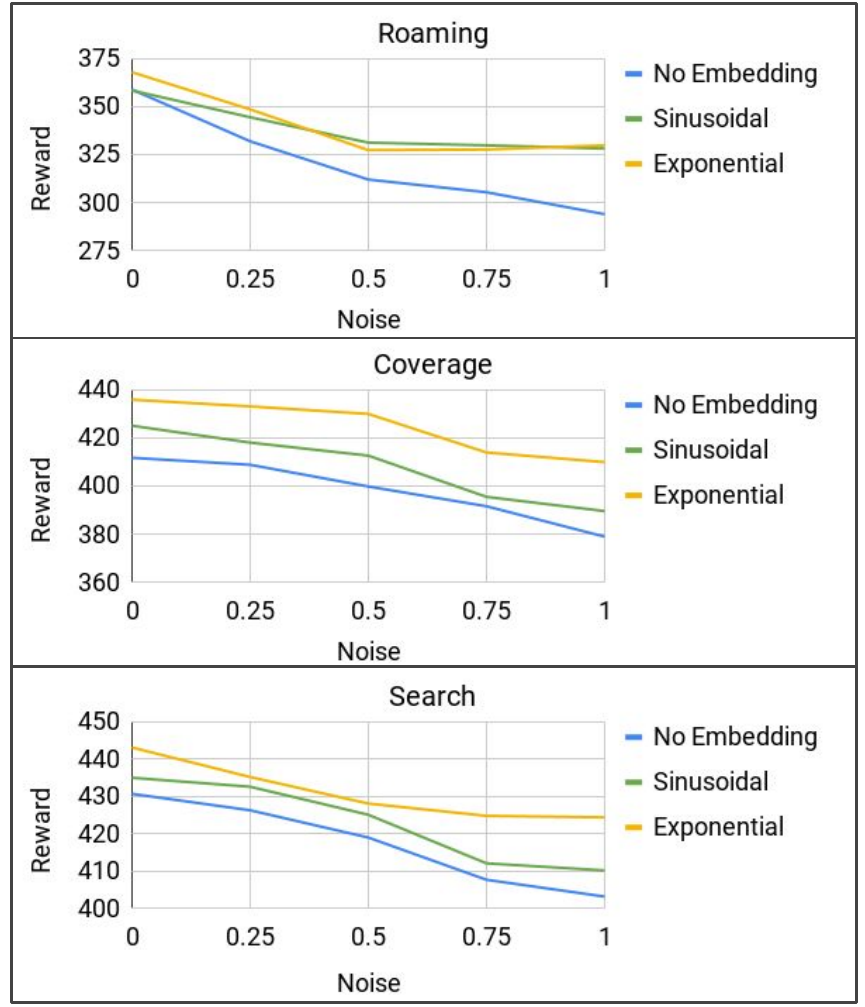}
    \caption{Robustness to noisy dynamics. We compare three positional embedding methods under noisy environment dynamics. The standard deviation of the noise is swept from 0.0 to 1.0.}
    \label{fig:robustness_to_noise}
\end{figure}

\section{Robustness to Noisy Dynamics}

In this section, we evaluate the robustness of our model to noisy environment dynamics. Instead of retrieving the ground truth poses $p_t$ from the environment, we estimate the pose using the action $a_t$. $a_t$ provides us translation and rotation of the agent \wrt the previous pose. Thus we can estimate the $\hat p_{t + 1}$ at each time step using $a_t$ and the previous estimation $\hat p_t$. When there is no noise, $\hat p_t$ is equivalent to $p_t$. With the Gaussian noise added at each time step, the noise added to $\hat p_t$ will be a Gaussian process. Therefore, when computing the observation embedding using the relative poses, recent steps suffer less from the noisy dynamics.

In our design of SMT, we use a positional embedding of the time step similar to \cite{Vaswani2017AttentionIA}, but with exponential functions instead of sinusoidal functions. The positional embedding provides temporal information of each time step for the policy. Sinusoidal function is periodic and provides only relative temporal information. In contrast, the exponential function is monotonic and represents how recent each time step is. In the long-horizon tasks we are interested in, we believe relative temporal information is not sufficient for the agent to understand long-term dependencies. 

To validate this assumption, we compare the exponential embedding with the two baselines. \textbf{No embedding} does not embed the positional embedding of the time step. \textbf{Sinusoidal} uses the same sinusoidal embedding function as in \cite{Vaswani2017AttentionIA}. We sweep the standard deviation of the noise from $0.0$ to $1.0$ and evaluates the average rewards on the validation set. In practice, we found the temporal information not only improves the performance given clean observations, but also helps leverage the noisy environment dynamics across time.  As shown in Fig.~\ref{fig:robustness_to_noise}, the average rewards decrease with more noises in dynamics. Sinusoidal and exponential embeddings both mitigate the performance drop. In the roaming task, the two embedding methods have comparable effects. While in coverage and search, exponential embedding has the superior performance.

\section{More Visualization}

\begin{figure*}[!t]
    \centering
    \includegraphics[width=\textwidth]{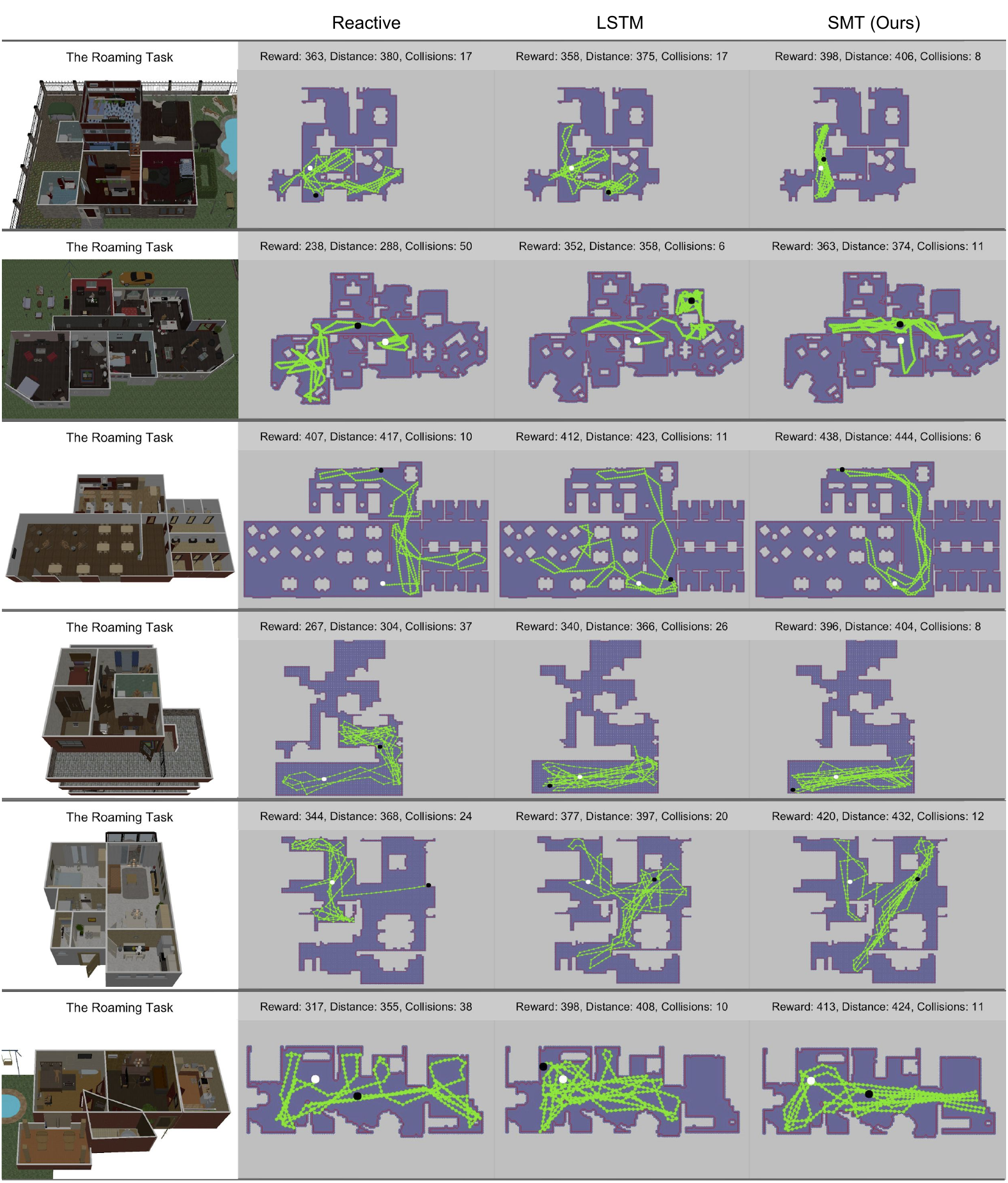}
    \caption{Visualization of the agent behaviors in the Roaming Task.}
    \label{fig:qulitative_roaming}
\end{figure*}

\begin{figure*}[!t]
    \centering
    \includegraphics[width=\textwidth]{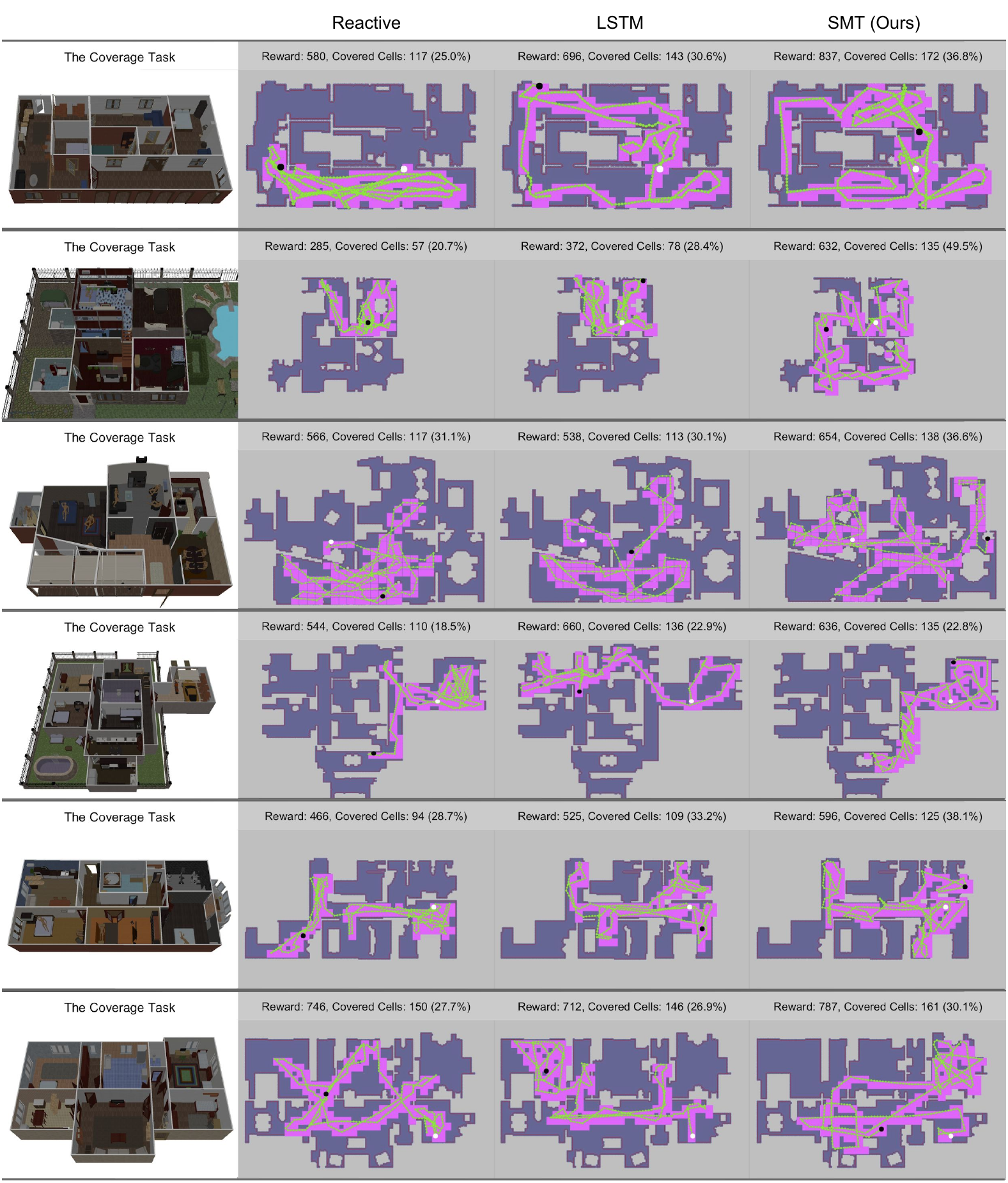}
    \caption{Visualization of the agent behaviors in the Coverage Task.}
    \label{fig:qulitative_coverage}
\end{figure*}

\begin{figure*}[!t]
    \centering
    \includegraphics[width=\textwidth]{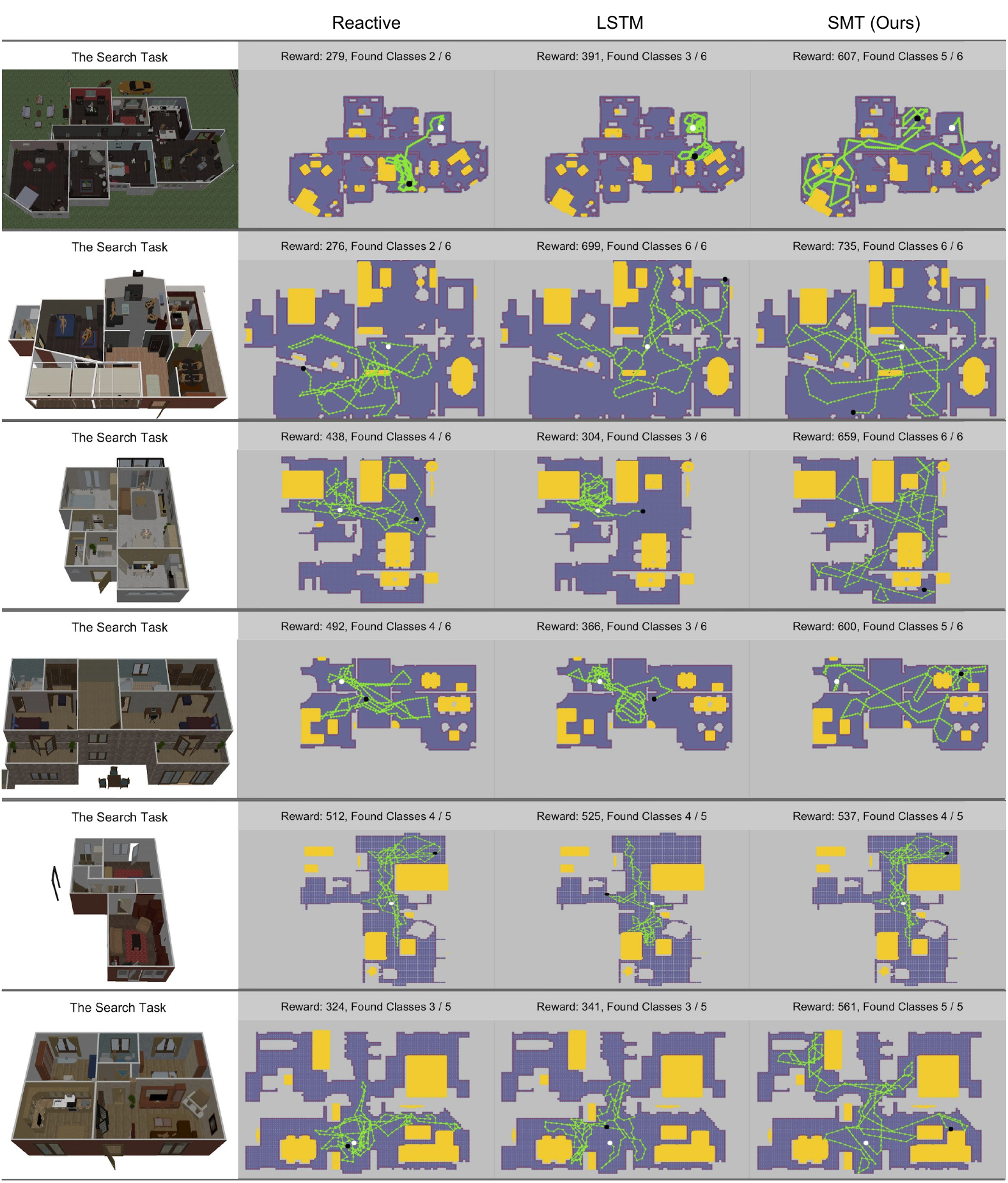}
    \caption{Visualization of the agent behaviors in the Search Task.}
    \label{fig:qulitative_search}
\end{figure*}

We present more visualization of the agent behaviors for roaming in Fig.~\ref{fig:qulitative_roaming}, for coverage in Fig.~\ref{fig:qulitative_coverage} and for search in Fig.~\ref{fig:qulitative_search}. As in the main paper, we visualize the trajectories from the top-down view as green curves, with white and black dots indicating the starting and ending points. Navigable area are masked in dark purple with red lines as the collision boundaries. In the coverage task (Fig.~\ref{fig:qulitative_coverage}), we mark the covered cells in pink. In the search task (Fig.~\ref{fig:qulitative_search}), we mark target objects with yellow masks. These figures demonstrate similar behaviors as analyzed in the main paper. The same reactive and LSTM baselines as in the main paper are used to compare with the proposed SMT policy.

\end{document}